\documentclass[11pt]{article}

\usepackage[]{EMNLP2022}

\usepackage{times}
\usepackage{latexsym}
\usepackage[T1]{fontenc}

\usepackage[utf8]{inputenc}

\usepackage{microtype}

\usepackage{inconsolata}

\usepackage{graphicx}

\usepackage{subcaption}

\usepackage{amsmath}

\title{Analysis of Levenshtein Transformer's Decoder and Its Variants}

\author{Ruiyang Zhou \\
  Télécom Paris \\}

\begin{document}
\maketitle
\begin{abstract}
Levenshtein transformer (LevT) is a non-autoregressive machine translation model with high decoding efficiency and comparable translation quality in terms of bleu score, due to its parallel decoding and iterative refinement procedure. Are there any deficiencies of its translations and what improvements could be made? In this report, we focus on LevT's decoder and analyse the decoding results length, subword generation, and deletion module's capability. We hope to identify weaknesses of the decoder for future improvements.

We also compare translations of the original LevT, knowledge-distilled LevT, LevT with translation memory, and the KD-LevT with translation memory to see how KD and translation memory can help.
\end{abstract}

\section{Introduction}
Classical neural machine translation models are autoregressive with good translation quality but low efficiency. The recently proposed non-autoregressive model \cite{gu2017non} alleviates the time-consuming problem. Its high efficiency comes from the parallel decoding structure, but as a side effect, parallelism reduces dependency between words and thus causes the multi-modality problem. 

Levenshtein Transformer (LevT) \cite{DBLP:journals/corr/abs-1905-11006} is proposed to fix problems caused by the lack of dependency between words. It allows modification of the parallel decoding result through an iterative refinement procedure. LevT achieves comparable results to autoregressive models. 

Sequence-level knowledge-distillation is another method to improve NAT model's performance \cite{kim2016sequence}. This method uses an autoregressive teacher to train a NAT student and can largely improve the result. 

In this report, we address the following research questions: How is the quality of original LevT translations? How are they different from knowledge-distilled LevT results? Are there any deficiencies in the decoding procedure? 

To answer these questions, we analyse the translation from three aspects: length prediction, subword/complete word generation, and deletion step. 

Our analysis shows that the (1) original LevT model predicts too short length in the beginning which leads to short final generations; (2) both original and knowledge-distilled LevT encounter a drop of performance when decoding from scratch; (3) both of them also make bad predictions on subwords in terms of number and quality; (4) both of them detect only grammatical errors and do not look at the source during deletion.~\footnote{During the internship in LISN with Jitao Xu and François Yvon.}

\section{Background}
\subsection{Data}
\begin{table}[ht]
\centering
\resizebox{\columnwidth}{!}{
  \begin{tabular}{l|ccc}
  \hline
  Domain & Raw & $\operatorname{sim}>0.6$ & $\operatorname{sim}\in[0.4, 0.6]$ \\
  \hline
  ECB & 195,956 &   51.73\% & 14.06\% \\
  EMEA & 373,235 &  65.68\% & 12.65\% \\
  Epps & 2,009,489 & 10.12\% & 25.30\% \\
  GNOME & 55,391 &  39.31\% & 11.06\% \\
  JRC & 503,437 &   50.87\% & 16.67\% \\
  KDE & 180,254 &   36.00\% & 10.81\% \\
  News & 151,423 &   2.12\% & 9.65\% \\
  PHP & 16,020 &    34.93\% & 12.38\% \\
  TED & 159,248 &   11.90\% & 26.64\% \\
  Ubuntu & 9,314 &  20.32\% & 8.26\% \\
  Wiki & 803,704 &  19.87\% & 17.32\% \\
  \hline
  Total & 4,457,471 & 24.27\% & 20.00\% \\
  \hline
  \end{tabular}}
\caption{Dataset statistics, with ratios of sentences with at least one TM match for various similarity ranges.}\label{tab:data-stat}
\end{table}

We use a multi-domain corpus with $11$ domains for English-French direction, collected from OPUS: documents from the European Central Bank (ECB); from the European Medicines Agency (EMEA); Proceedings of the European Parliament (Epps); legislative texts of the European Union (JRC); News Commentaries (News); TED talk subtitles (TED); parallel sentences extracted from Wikipedia (Wiki); localization files (GNOME, KDE and Ubuntu) and manuals (PHP). All these data are deduplicated prior to training. Table~\ref{tab:data-stat} reports statistics of the ratio of TM matches for various similarity ranges; these ratios vary greatly across domains. We tokenize all data using Moses and build a shared source-target vocabulary with $32$K BPE learned with \texttt{subword-nmt}.\footnote{\url{https://github.com/rsennrich/subword-nmt}}

\begin{figure*}[ht]
\centering
\includegraphics[width=\textwidth]{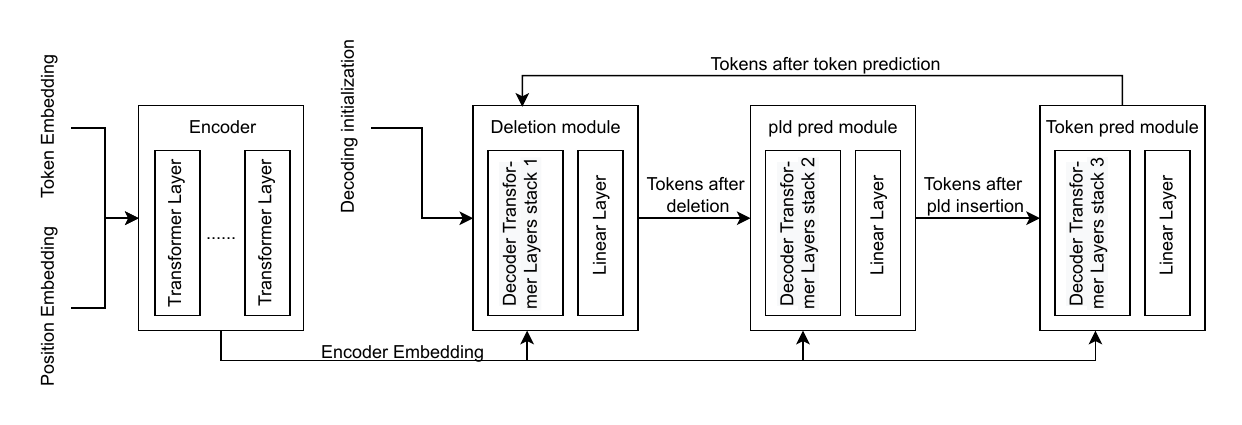}
\caption{LevT structure when inferencing.}
\label{fig:levt-infer}
\end{figure*}

\begin{figure*}[ht]
\centering
\includegraphics[width=\textwidth]{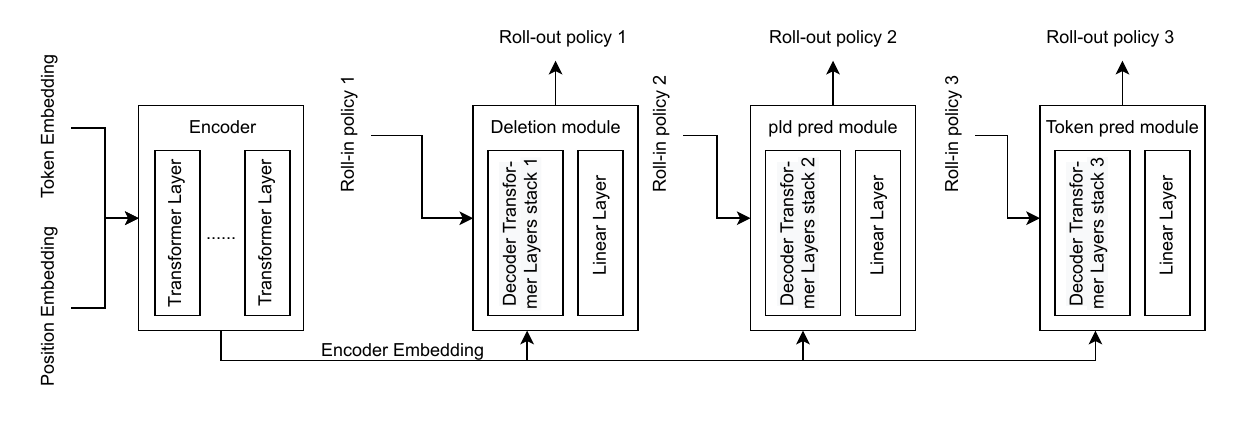}
\caption{LevT structure when training.}
\label{fig:levt-train}
\end{figure*}

\subsection{LevT model}
Levenshtein Transformer (LevT) is a non autoregressive translation model with iterative refinement \cite{DBLP:journals/corr/abs-1905-11006}. It has a Transformer-based encoder \cite{DBLP:journals/corr/VaswaniSPUJGKP17} and also a Transformer-based decoder with three modules: deletion module, placeholder (pld) length prediction module, token prediction module. 

As shown in Figure~\ref{fig:levt-infer}, the encoder encodes both tokens and positions. Each module of the decoder contains a transformer base (can choose to share or not, here we choose to share so that the three modules use the same transformer decoder base) and a linear layer.

The decoding process is iterative: starts from the initial sentence (empty by default), it deletes (if empty, then delete nothing), predicts the length of needed tokens and inserts the corresponding number of pld tokens, then predicts words to replace pld tokens, finally iterates again from the deletion step. The refinement process ends if the sentence does not change between two iterations, or reaches a maximum number of rounds.

Deletion prediction module performs a classification task: for each token, it predicts scores for $0$ (not to delete) and $1$ (to delete) and keeps the higher one. Pld prediction module performs a classification task: between each two consecutive tokens, it gives scores to the length of insertion (from $0$ to $255$) and keeps the highest one. Token prediction module also performs a classification task: at each pld token position, it chooses the word of highest score from the dictionary to replace the pld token.

\subsection{LevT training}
As shown in Figure~\ref{fig:levt-train}, we set different input and label generation method for each module of the decoder. All labels are generated by minimizing the Levenshtein distance, all modules use cross-entropy loss, and the total loss is the sum of three losses. The descriptions are based on the published LevT code in Fairseq. \footnote{\url{https://github.com/facebookresearch/fairseq/tree/main/examples/nonautoregressive_translation}}

For pld length prediction module, input is generated by dropping each word of the reference target sentence according to a uniform distribution between $0$ and $1$. For token prediction module, input is generated by masking the same words that are dropped previously (i.e. reference pld insertion result). 

The reference label of pld length prediction and token prediction can be produced at once. It is obtained by calculating the best insertion strategy (in terms of Levenshtein distance) between dropped sentence and reference target sentence. We can also get labels by keeping track of which words are dropped. LevT then uses the reference label to compute cross-entropy loss.

For deletion prediction module, its input is the generated sentence after token prediction module. Instead of choosing the $argmax$ prediction, we sample from all predictions according to the weight $softmax(score)$ to increase training data variety. The reference label is also obtained from the lowest editing distance, and the model also uses cross-entropy loss. 

To summarize:
\begin{itemize}
    \item Roll-in policy $1$: output of token prediction module.
    \item Roll-in policy $2$: drop each token of the reference translation according to a uniform distribution $U[0,1]$.
    \item Roll-in policy $3$: mask by [pld] the same tokens that are dropped in policy $2$.
    \item Roll-out policy $1$ / $2$\&$3$: best editing actions from policy $1$ / $2$ to reference translation.
\end{itemize}

\subsection{Sequence-level knowledge-distillation}
 Sequence-level knowledge-distillation is a widely used method for training non-autoregressive translation models \cite{kim2016sequence}, because it largely improves NAT model's performance. But in what way the generations are improved and are there still deficiencies? In the following sections, we compare the generations of knowledge-distilled LevT to that of the original LevT.
 
 Sequence-level knowledge-distillation uses a teacher model (usually a autoregressive translation model) to generate translation on training set, then takes its generations as reference sentences to train the student model. The translation from teacher model is produced by beam search, and is used by the student model to compute cross-entropy loss. 

Knowledge-distillation method can alleviate multi-modality problem on NAT models. According to \cite{zhou2019understanding}, it reduces the complexity of training set while keeping the faithfulness of reference translations. 

\subsection{LevT with translation memory}
Training LevT using translation memory is another method to improve the performance. When training the Levenshtein transformer, we give it existing translations as decoding initialization, so that the model learns to modify given sentences. We also compare its generations to the original LevT.

\section{Analysis of length}
\subsection{Original LevT produces short translations}
The process of iterative refinement is: deletion 1 -> \emph{pld insertion 1} -> token prediction 1 -> \emph{deletion 2} -> \emph{pld insertion 2} -> ... -> \emph{final translation}.
We compute average sentence length (BPE-token level) of the first two and the last iterations (steps in italic format).
We also compute length without stop word to evaluate translation quality.

\begin{table}[ht]
\centering
\resizebox{\columnwidth}{!}{
\begin{tabular}{ccccccc}
\hline
\textbf{length}    & \multicolumn{2}{c}{\textbf{Orig}} & \multicolumn{2}{c}{\textbf{KD}} & \multicolumn{2}{c}{\textbf{TM}} \\
                   & len      & len\textbackslash{}    & len     & len\textbackslash{}   & len     & len\textbackslash{}   \\ \hline
\textbf{ref}       & 19.75    & 12.41                  & 19.75   & 12.41                 & 19.75   & 12.41                 \\
\textbf{pld 1} & 15.46    & 9.79                   & 18.65   & 12.18                  & 15.82   & 10.20                 \\
\textbf{del 1}     & 12.58    & 7.79                   & 17.08   & 10.91                  & 13.09   & 8.16                  \\
\textbf{pld 2} & 15.08    & 9.25                   & 18.66   & 11.71                  & 15.66   & 9.65                  \\
\textbf{final}     & 15.33    & 9.38                   & 18.75   & 11.75                 & 15.94   & 9.80                  \\ \hline
\end{tabular}}
\caption{Translation length in each iteration (bpe level). 'len\textbackslash{}' means length without stop word.}
\label{tab:length}
\end{table}

\paragraph{Observation} From Table~\ref{tab:length}, we find that 
(1) Orig-LevT has short bpe-level length and bad word-level BP score. It does not delete many and not generates many. (2) KD-LevT does not have length problem. It deletes more and generates more in each iteration.

\subsection{All models do not have duplication and invalid word problem}
\begin{table}[ht]
\centering
\resizebox{\columnwidth}{!}{
\begin{tabular}{ccccccc}
\hline
\textbf{length}    & \multicolumn{2}{c}{\textbf{Orig}} & \multicolumn{2}{c}{\textbf{KD}} & \multicolumn{2}{c}{\textbf{TM}} \\
                   & num      & num\textbackslash{}    & num     & num\textbackslash{}   & num     & num\textbackslash{}   \\ \hline
\textbf{ref}       & 0    & 0    & 0    & 0    & 0    & 0    \\
\textbf{tok 1} & 2.06 & 1.41 & 1.26 & 1.07 & 1.84 & 1.41 \\
\textbf{del 1}     & 0.07 & 0.05 & 0.06 & 0.06 & 0.06 & 0.04 \\
\textbf{tok 2} & 0.05 & 0.06 & 0.05 & 0.06 & 0.04 & 0.05 \\
\textbf{final}     & 0.05 & 0.05 & 0.06 & 0.07 & 0.05 & 0.06 \\ \hline
\end{tabular}}
\caption{Number of duplicated token (bpe level) with and without stop word.}
\label{tab:duplication}
\end{table}

\begin{table}[ht]
\centering
\resizebox{\columnwidth}{!}{
\begin{tabular}{cccc}
\hline
\textbf{invalid word}      & \textbf{Orig} & \textbf{KD} & \textbf{TM} \\ \hline
\textbf{num of word/total} & 750 / 140106    & 844 / 167886  & 943 / 148988  \\
\textbf{ratio of sent}     & 0.0605        & 0.0608      & 0.0759      \\ \hline
\end{tabular}}
\caption{Number of invalid word (word level) and of sentences containing invalid word.}
\label{tab:invalid word}
\end{table}

Since the model is non-autoregressive, it is possible that the lack of word dependency during generation could lead to duplication or invalid word problem. For duplication, we count duplicated tokens (bpe level) for sentences with and without stop word in each iteration, e.g. the sentence 'a b b b c' has two duplication. For invalid word, we count the total number (word level) and the number of sentences containing invalid word in final translations. 

\paragraph{Observation}From Table~\ref{tab:duplication} and~\ref{tab:invalid word}, (1) Although model has duplication at first generation, it is able to correct them in the end. Each sentence contains only 0.05 duplicated tokens in average. (2) Three models have similar proportion of invalid word, it is not a peculiar problem.

\subsection{Try other length choice}\label{5len}
During decoding, length prediction is set as a classification task. The model predicts scores for each length choice (0-255) and chooses the maximum one. Here, we try two other ways of choosing the length prediction. (1) During first iteration, we keep top five length predictions and generate five translations. For each translation, the refinement process is as usual. (2) During second iteration, we randomly choose length predictions according to their score, i.e. multinomial distribution whose weight is softmax(score). 

We also count the duplication and invalid word in final translation. Both models do not have this problem, the proportions are always around 0.05 and 0.07.

\begin{table}[ht]
\centering
\resizebox{\columnwidth}{!}{
\begin{tabular}{cccccc}
\hline
\textbf{Orig-LevT}        & \textbf{1st}                                                                         & \textbf{2nd}                                                                         & \textbf{3rd}                                                                         & \textbf{4th}                                                                         & \textbf{5th}                                                                         \\ \hline
\textbf{iter num}       & 1.99                                                                                 & 2.24                                                                                 & 2.33                                                                                 & 2.42                                                                                 & 2.47                                                                                 \\
\textbf{pld 1 len}  & 15.48                                                                                & 15.94                                                                                & 16.33                                                                                & 16.74                                                                                & 16.85                                                                                \\
\textbf{pld 2 len}  & 15.08                                                                                & 15.81                                                                                & 16.09                                                                                & 16.39                                                                                & 16.44                                                                                \\
\textbf{last len}       & 16.33                                                                                & 17.1                                                                                 & 17.41                                                                                & 17.72                                                                                & 17.79                                                                                \\
\textbf{tok 1 bleu} & \begin{tabular}[c]{@{}c@{}}34.0568 \\ BP = 0.783\end{tabular} & \begin{tabular}[c]{@{}c@{}}33.3507 \\ BP = 0.805\end{tabular} & \begin{tabular}[c]{@{}c@{}}32.5583 \\ BP = 0.826\end{tabular} & \begin{tabular}[c]{@{}c@{}}31.9184 \\ BP = 0.849\end{tabular} & \begin{tabular}[c]{@{}c@{}}30.7359 \\ BP = 0.857\end{tabular} \\
\textbf{final bleu}     & \begin{tabular}[c]{@{}c@{}}45.4371 \\ BP = 0.774\end{tabular} & \begin{tabular}[c]{@{}c@{}}47.3946 \\ BP = 0.818\end{tabular} & \begin{tabular}[c]{@{}c@{}}47.9981 \\  BP = 0.835\end{tabular} & \begin{tabular}[c]{@{}c@{}}48.2303 \\ BP = 0.850\end{tabular}  & \begin{tabular}[c]{@{}c@{}}48.2863 \\ BP = 0.855\end{tabular} \\ \hline
\end{tabular}}
\caption{Five predictions on Orig-LevT.}
\label{tab:5len-orig}
\end{table}

\begin{table}[ht]
\centering
\resizebox{\columnwidth}{!}{
\begin{tabular}{cccccc}
\hline
\textbf{KD-LevT}        & \textbf{1st}                                                                         & \textbf{2nd}                                                                         & \textbf{3rd}                                                                         & \textbf{4th}                                                                         & \textbf{5th}                                                                         \\ \hline
\textbf{iter num}       & 1.54                                                                                 & 1.88                                                                                 & 2                                                                                    & 2.13                                                                                 & 2.21                                                                                 \\
\textbf{pld 1 len}  & 18.65                                                                                & 18.67                                                                                & 18.67                                                                                & 19.04                                                                                & 19.35                                                                                \\
\textbf{pld 2 len}  & 18.65                                                                                & 18.76                                                                                & 18.84                                                                                & 19.18                                                                                & 19.53                                                                                \\
\textbf{final len}      & 19.75                                                                                & 19.88                                                                                & 19.96                                                                                & 20.29                                                                                & 20.57                                                                                \\
\textbf{tok 1 bleu} & \begin{tabular}[c]{@{}c@{}}46.2238 \\ BP = 0.945\end{tabular} & \begin{tabular}[c]{@{}c@{}}42.6489 \\ BP = 0.946\end{tabular} & \begin{tabular}[c]{@{}c@{}}40.5684 \\ BP = 0.945\end{tabular} & \begin{tabular}[c]{@{}c@{}}38.9331 \\ BP = 0.965\end{tabular} & \begin{tabular}[c]{@{}c@{}}37.8941 \\ BP = 0.981\end{tabular} \\
\textbf{final bleu}     & \begin{tabular}[c]{@{}c@{}}53.6718 \\ 73.7/59.7/51.1/44.7 \\ BP = 0.953\end{tabular} & \begin{tabular}[c]{@{}c@{}}52.6441 \\ BP = 0.959\end{tabular} & \begin{tabular}[c]{@{}c@{}}52.6405 \\ BP = 0.964\end{tabular} & \begin{tabular}[c]{@{}c@{}}52.0828 \\ BP = 0.980\end{tabular} & \begin{tabular}[c]{@{}c@{}}51.6052 \\ BP = 0.994\end{tabular} \\ \hline
\end{tabular}}
\caption{Five predictions on KD-LevT.}
\label{tab:5len-kd}
\end{table}

\begin{table}[ht]
\centering
\resizebox{\columnwidth}{!}{
\begin{tabular}{cccccc}
\hline
\textbf{Orig-LevT}      & \textbf{1st}                                                                         & \textbf{2nd}                                                                         & \textbf{3rd}                                                                         & \textbf{4th}                                                                         & \textbf{5th}                                                                         \\ \hline
\textbf{iter num}       & 2.63                                                                                 & 2.71                                                                                 & 2.75                                                                                 & 2.8                                                                                  & 2.84                                                                                 \\
\textbf{pld 1 len}  & 15.48                                                                                & 15.94                                                                                & 16.34                                                                                & 16.72                                                                                & 16.87                                                                                \\
\textbf{pld 2 len}  & 23.55                                                                                & 23.76                                                                                & 23.73                                                                                & 23.69                                                                                & 23.29                                                                                \\
\textbf{final len}      & 18.99                                                                                & 19.24                                                                                & 19.52                                                                                & 19.58                                                                                & 19.55                                                                                \\
\textbf{tok 1 bleu} & \begin{tabular}[c]{@{}c@{}}34.0744 \\ BP = 0.784\end{tabular} & \begin{tabular}[c]{@{}c@{}}33.3143 \\ BP = 0.805\end{tabular} & \begin{tabular}[c]{@{}c@{}}32.5499 \\ BP = 0.826\end{tabular} & \begin{tabular}[c]{@{}c@{}}31.9420 \\ BP = 0.848\end{tabular} & \begin{tabular}[c]{@{}c@{}}30.7469 \\ BP = 0.858\end{tabular} \\
\textbf{final bleu}     & \begin{tabular}[c]{@{}c@{}}47.9868 \\ BP = 0.930\end{tabular} & \begin{tabular}[c]{@{}c@{}}48.5947 \\ BP = 0.941\end{tabular} & \begin{tabular}[c]{@{}c@{}}49.1004 \\ BP = 0.954\end{tabular}   & \begin{tabular}[c]{@{}c@{}}48.9312 \\ BP = 0.954\end{tabular} & \begin{tabular}[c]{@{}c@{}}48.9097 \\ BP = 0.952\end{tabular} \\ \hline
\end{tabular}}
\caption{Five predictions plus random choice on Orig-LevT.}
\label{tab:5len+random-orig}
\end{table}

\paragraph{Observation} (1) From Table~\ref{tab:5len-orig}, among the top five length predictions, Orig-LevT assigns lower score to longer length. It brings better BP score and thus better bleu score, though with a small drop of ngram precision. (2) From Table~\ref{tab:5len-kd}, for KD-LevT, as its BP score is already good, other length choices only lead to worse precision and thus worse bleu score. (3) By comparing the two, we find that Orig-LevT has bad length prediction. Choosing worse-scored but longer length prediction can alleviate this problem and increase final bleu score. (4) From Table~\ref{tab:5len+random-orig}, choosing worse length prediction needs more refinement iterations after first token insertion. For Orig-LevT, it also deletes more when bad length.

\subsection{Use external length predictor}
As in \ref{5len}, we find that length prediction in first iteration is crucial. Here, we try four different external length prediction methods (bpe level) and let the model use this external value as its length in first iteration. Four methods are: (1) source sentence length. (2) $srclen * ratio$ , where the average ratio is computed from training set $1.06$. (3) linear regression trained on training set, $coef=0.726, \ intercept=10.2, \ R^2=0.754$. (4) reference length. 

\begin{table}[ht]
\centering
\resizebox{\columnwidth}{!}{
\begin{tabular}{ccccc}
\hline
\textbf{Orig-LevT}          & \textbf{srclen}                                                                      & \textbf{srclen * ratio}                                                              & \textbf{linear reg}                                                                  & \textbf{tgtlen}                                                                      \\ \hline
\textbf{iter num}           & 2.23                                                                                 & 2.24                                                                                 & 2.64                                                                                 & 2.1                                                                                  \\
\textbf{external pld 1 len} & 17.45                                                                                & 18.46                                                                                & 22.73                                                                                & 19.24                                                                                \\
\textbf{pld 2 len}          & 17.43                                                                                & 17.88                                                                                & 18.64                                                                                & 17.84                                                                                \\
\textbf{final len}          & 18.74                                                                                & 19.17                                                                                & 19.88                                                                                & 19.07                                                                                \\
\textbf{tok 1 bleu}         & \begin{tabular}[c]{@{}c@{}}33.6824  \\ BP = 0.886\end{tabular} & \begin{tabular}[c]{@{}c@{}}32.9922  \\ BP = 0.674\end{tabular} & \begin{tabular}[c]{@{}c@{}}27.4284  \\ BP = 1.000\end{tabular} & \begin{tabular}[c]{@{}c@{}}41.1424 \\ BP = 0.990\end{tabular} \\
\textbf{final bleu}         & \begin{tabular}[c]{@{}c@{}}51.1310 \\ BP = 0.907\end{tabular} & \begin{tabular}[c]{@{}c@{}}52.0311 \\ BP = 0.929\end{tabular} & \begin{tabular}[c]{@{}c@{}}50.1142 \\ BP = 0.962\end{tabular} & \begin{tabular}[c]{@{}c@{}}55.3332 \\ BP = 0.925\end{tabular} \\ \hline
\end{tabular}}
\caption{Try external length predictor for first pld insertion length on Orig-LevT.}
\label{tab:ext-len-orig}
\end{table}

\begin{table}[ht]
\centering
\resizebox{\columnwidth}{!}{
\begin{tabular}{ccccc}
\hline
\textbf{KD-LevT}            & \textbf{srclen}                                                                      & \textbf{srclen * ratio}                                                              & \textbf{linear reg}                                                                  & \textbf{tgtlen}                                                                      \\ \hline
\textbf{iter num}           & 1.89                                                                                 & 1.92                                                                                 & 2.54                                                                                 & 1.79                                                                                 \\
\textbf{external pld 1 len} & 17.45                                                                                & 18.46                                                                                & 22.73                                                                                & 19.24                                                                                \\
\textbf{pld 2 len}          & 18.37                                                                                & 18.85                                                                                & 21.61                                                                                & 19.14                                                                                \\
\textbf{final len}          & 19.55                                                                                & 20                                                                                   & 22.69                                                                                & 20.26                                                                                \\
\textbf{tok 1 bleu}         & \begin{tabular}[c]{@{}c@{}}38.1862 \\ BP = 0.879\end{tabular} & \begin{tabular}[c]{@{}c@{}}38.7210 \\ BP = 0.798\end{tabular} & \begin{tabular}[c]{@{}c@{}}32.3401 \\ BP = 1.000\end{tabular} & \begin{tabular}[c]{@{}c@{}}46.6037 \\ BP = 0.976\end{tabular} \\
\textbf{final bleu}         & \begin{tabular}[c]{@{}c@{}}52.2283 \\ BP = 0.946\end{tabular} & \begin{tabular}[c]{@{}c@{}}52.7395 \\ BP = 0.967\end{tabular} & \begin{tabular}[c]{@{}c@{}}45.8535 \\ BP = 1.000\end{tabular} & \begin{tabular}[c]{@{}c@{}}55.5059 \\ BP = 0.980\end{tabular} \\ \hline
\end{tabular}}
\caption{Try external length predictor for first pld insertion length on KD-LevT.}
\label{tab:ext-len-kd}
\end{table}

\begin{table}[ht]
\centering
\resizebox{\columnwidth}{!}{
\begin{tabular}{ccccc}
\hline
\textbf{TM-LevT}            & \textbf{srclen}                                                                      & \textbf{srclen * ratio}                                                              & \textbf{linear reg}                                                                  & \textbf{tgtlen}                                                                      \\ \hline
\textbf{iter num}           & 2                                                                                    & 2.03                                                                                 & 2.4                                                                                  & 1.86                                                                                 \\
\textbf{external pld 1 len} & 17.45                                                                                & 18.46                                                                                & 22.73                                                                                & 19.24                                                                                \\
\textbf{pld 2 len}          & 17.74                                                                                & 18                                                                                   & 18.45                                                                                & 17.91                                                                                \\
\textbf{final len}          & 18.81                                                                                & 19.01                                                                                & 19.58                                                                                & 18.94                                                                                \\
\textbf{tok 1 bleu}         & \begin{tabular}[c]{@{}c@{}}34.3025 \\ BP = 0.888\end{tabular} & \begin{tabular}[c]{@{}c@{}}32.6684 \\ BP = 0.646\end{tabular} & \begin{tabular}[c]{@{}c@{}}28.0365 \\ BP = 1.000\end{tabular} & \begin{tabular}[c]{@{}c@{}}42.7462 \\ BP = 0.990\end{tabular} \\
\textbf{final bleu}         & \begin{tabular}[c]{@{}c@{}}51.8304 \\ BP = 0.905\end{tabular} & \begin{tabular}[c]{@{}c@{}}52.3598 \\ BP = 0.916\end{tabular} & \begin{tabular}[c]{@{}c@{}}51.8092 \\ BP = 0.942\end{tabular} & \begin{tabular}[c]{@{}c@{}}55.2741 \\ BP = 0.914\end{tabular} \\ \hline
\end{tabular}}
\caption{Try external length predictor for first pld insertion length on TM-LevT.}
\label{tab:ext-len-tm}
\end{table}

\paragraph{Observation} From Table~\ref{tab:ext-len-orig}, \ref{tab:ext-len-kd}, \ref{tab:ext-len-tm}, we find that (1) in average, linear regression gives worse length prediction than simply apply a ratio (linear regression without intercept). (2) for all three models, variations of final prediction length are all smaller than initial length, showing that models tend to end at correct length. (3) If we replace its length prediction in the first iteration by the reference sentence length, all three models achieve similar good final bleu score, showing that the first length prediction is crucial.

\section{Analysis of subword VS complete word}
\subsection{Original LevT generates insufficient subwords}
The model generates BPE level translation, so that a generated token could be a subword token or a complete word token, depending on its frequency in training data. The difficulty of generating several subword tokens and generating one complete word may be different, so we count the proportion of subword tokens among all generated tokens (bpe level) to verify if it is a real problem. For example, 'a@@ b@@ c' are three subwords.

\begin{table}[ht]
\centering
\resizebox{\columnwidth}{!}{
\begin{tabular}{ccccc}
\hline
\textbf{}                                                                        & \textbf{ref} & \textbf{Orig} & \textbf{KD} & \textbf{TM} \\ \hline
\textbf{\begin{tabular}[c]{@{}c@{}}sub ratio \\ (=sub num/tok num)\end{tabular}} & 0.251        & 0.204         & 0.232       & 0.227       \\
\textbf{sub num}                                                                 & 4.95         & 3.33          & 4.59        & 4.3         \\
\textbf{tok num}                                                                 & 19.75        & 16.34         & 19.74       & 18.94       \\ \hline
\end{tabular}}
\caption{Proportion of subword tokens among all generated tokens (bpe level).}
\label{tab:subword}
\end{table}

\paragraph{Observation} From Table~\ref{tab:subword}, (1) All three models generate less subword tokens than real reference translation. (2) KD-LevT has better subword generation. 

\subsection{Test model's ability to predict subword token and full word token}
Token generation consists of two steps: pld length prediction and token prediction. In order to test both abilities, we give the decoder an incomplete sentence as decoding initialization, let it predict pld length and tokens, then measure the predictions. There are two types of incomplete sentence: missing only subword tokens and missing only complete word tokens. 

To produce these sentences and labels, we take the reference target sentence and randomly delete $5\%, 10\%, 15\%, 20\%, 25\%$ tokens that are subword tokens or complete word tokens, then generate the correct pld prediction labels accordingly. The maximum ratio $25\%$ comes from the real percentage of subword tokens in reference translations. 

We also randomly delete $10\%, 20\% ... 100\%$ tokens of reference target translation as decoding initialization, regardless of subword or complete word. 

Method to compute pld prediction accuracy: $avg(pred == ref)$, e.g. pred = $0020$, ref = $0120$, accuracy = $1/4$. 

Method to count matched token number: between each two tokens of decoding initialization, there should be one pld length prediction and some token predictions; we count matched token number of this pld length prediction regardless of order, then average over the whole sentence. e.g. init = a d,  tgt = a b c d, pred = a c b d, match = 2. 

\begin{figure}[ht]
\centering
\begin{subfigure}{.48\columnwidth}
  \centering
  \includegraphics[width=\columnwidth]{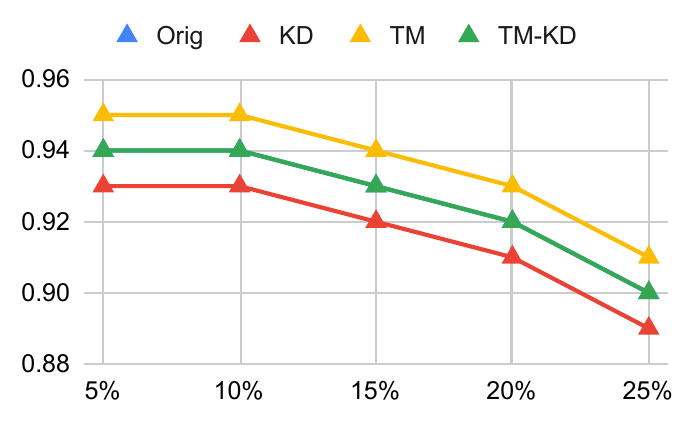}
  \caption{for subword token}
  \label{fig:pld-sub}
\end{subfigure}%
\begin{subfigure}{.48\columnwidth}
  \centering
  \includegraphics[width=\columnwidth]{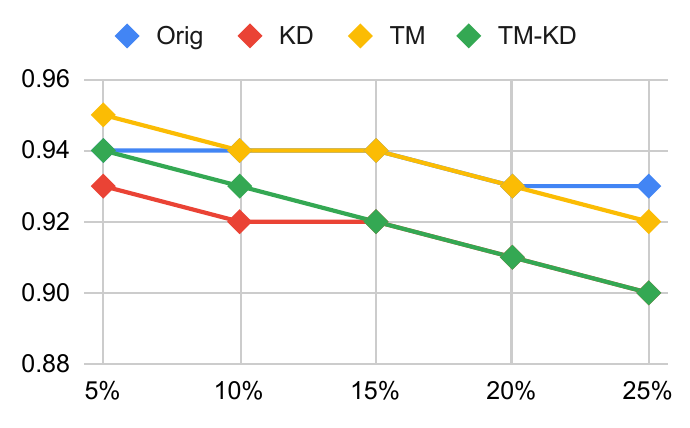}
  \caption{for full word token}
  \label{fig:pld-word}
\end{subfigure}
\caption{Pld prediction accuracy of four models}
\label{fig:pld}
\end{figure}

\begin{figure}[ht]
\centering
\begin{subfigure}{.48\columnwidth}
  \centering
  \includegraphics[width=\columnwidth]{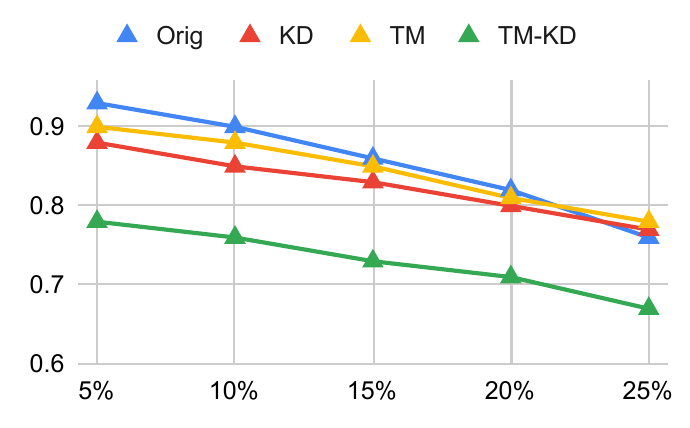}
  \caption{precision of subword}
  \label{fig:precision-sub}
\end{subfigure}%
\begin{subfigure}{.48\columnwidth}
  \centering
  \includegraphics[width=\columnwidth]{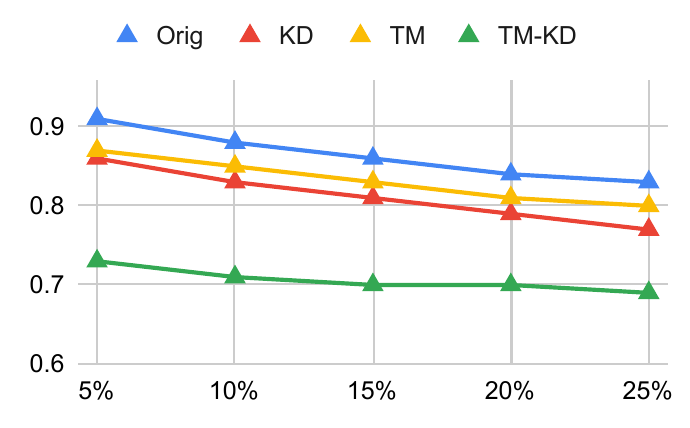}
  \caption{precision of full word}
  \label{fig:precision-word}
\end{subfigure}
\vskip\baselineskip
\begin{subfigure}{.48\columnwidth}
  \centering
  \includegraphics[width=\columnwidth]{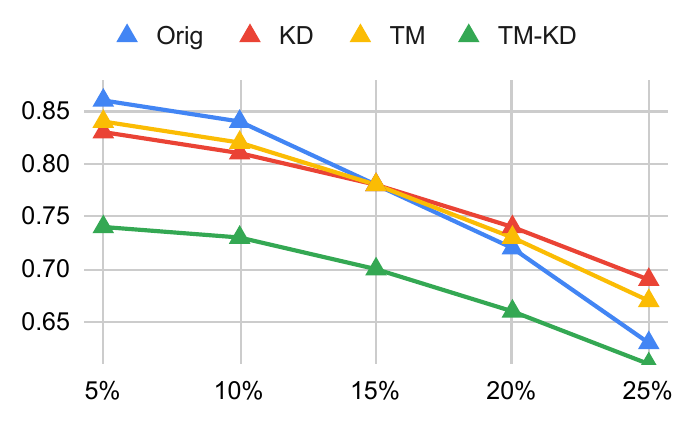}
  \caption{recall of subword}
  \label{fig:recall-sub}
\end{subfigure}%
\begin{subfigure}{.48\columnwidth}
  \centering
  \includegraphics[width=\columnwidth]{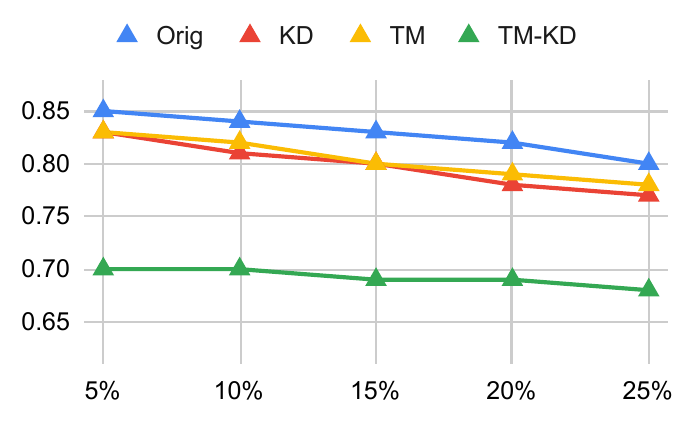}
  \caption{recall of full word}
  \label{fig:recall-word}
\end{subfigure}
\caption{Precision and recall of subword/complete word token prediction.}
\label{fig:sub/word}
\end{figure}

\begin{figure}[ht]
\centering
\begin{subfigure}{.48\columnwidth}
  \centering
  \includegraphics[width=\columnwidth]{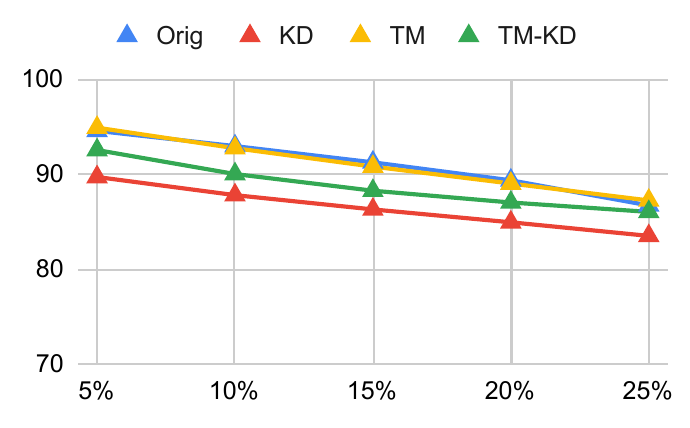}
  \caption{subword after one round}
  \label{fig:bleu-first-sub}
\end{subfigure}%
\begin{subfigure}{.48\columnwidth}
  \centering
  \includegraphics[width=\columnwidth]{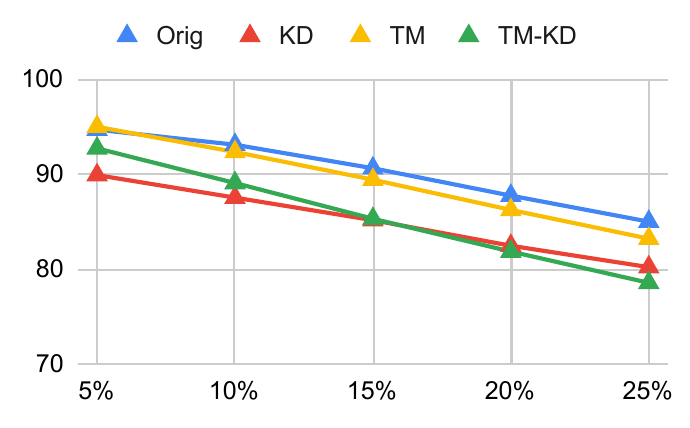}
  \caption{full word after one round}
  \label{fig:bleu-first-word}
\end{subfigure}
\vskip\baselineskip
\begin{subfigure}{.48\columnwidth}
  \centering
  \includegraphics[width=\columnwidth]{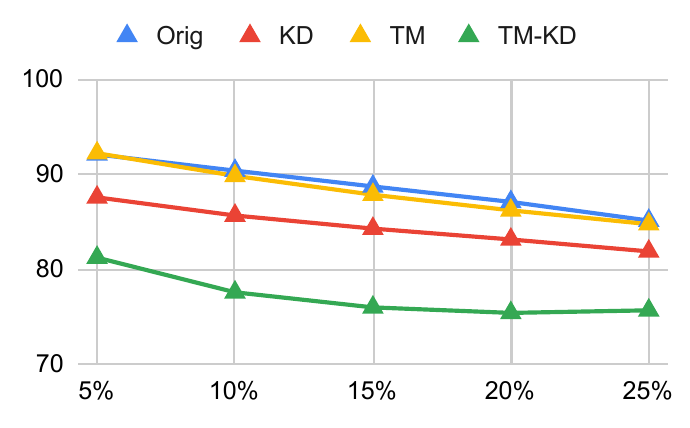}
  \caption{subword at last}
  \label{fig:bleu-last-sub}
\end{subfigure}%
\begin{subfigure}{.48\columnwidth}
  \centering
  \includegraphics[width=\columnwidth]{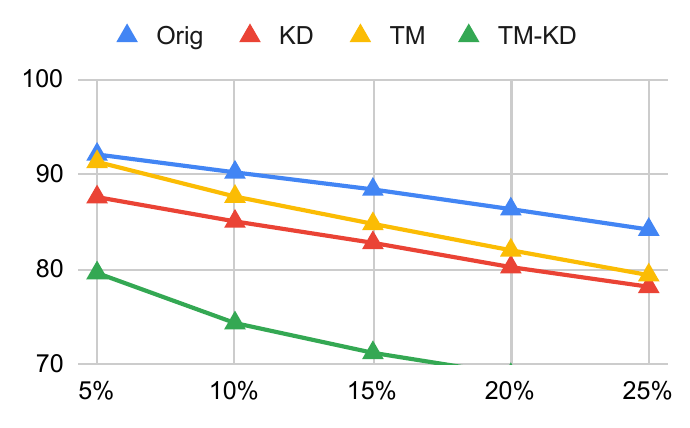}
  \caption{full word at last}
  \label{fig:bleu-last-word}
\end{subfigure}
\caption{Word level bleu score with decoding initialization missing only subword or only complete word.}
\label{fig:sub/word-bleu}
\end{figure}

\begin{figure}[ht]
\centering
\begin{subfigure}{.96\columnwidth}
  \centering
  \includegraphics[width=\columnwidth]{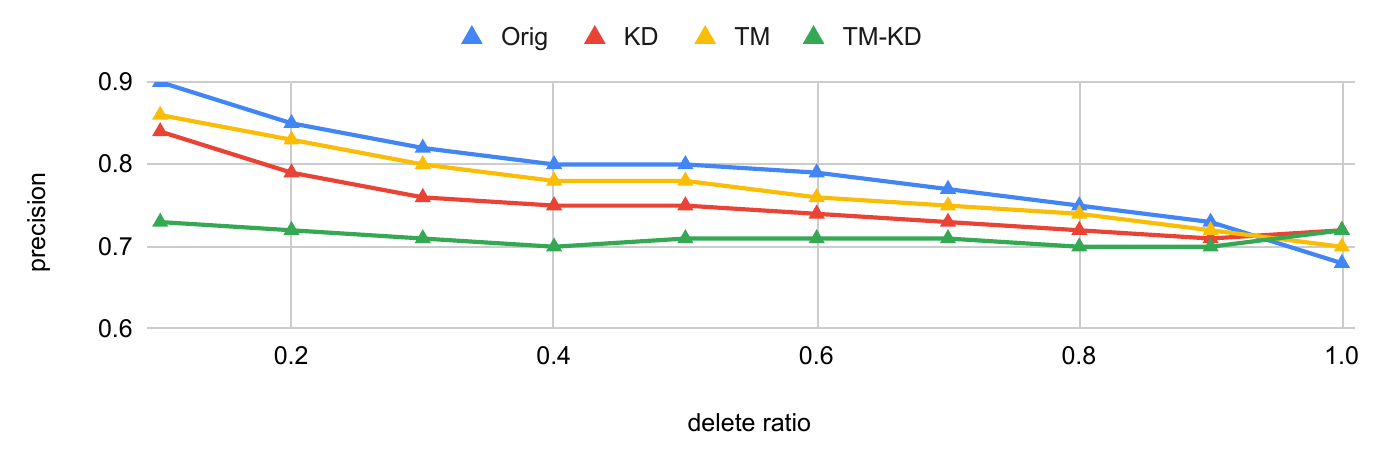}
  \caption{precision of random token}
  \label{fig:precision}
\end{subfigure}%
\vskip\baselineskip
\begin{subfigure}{.96\columnwidth}
  \centering
  \includegraphics[width=\columnwidth]{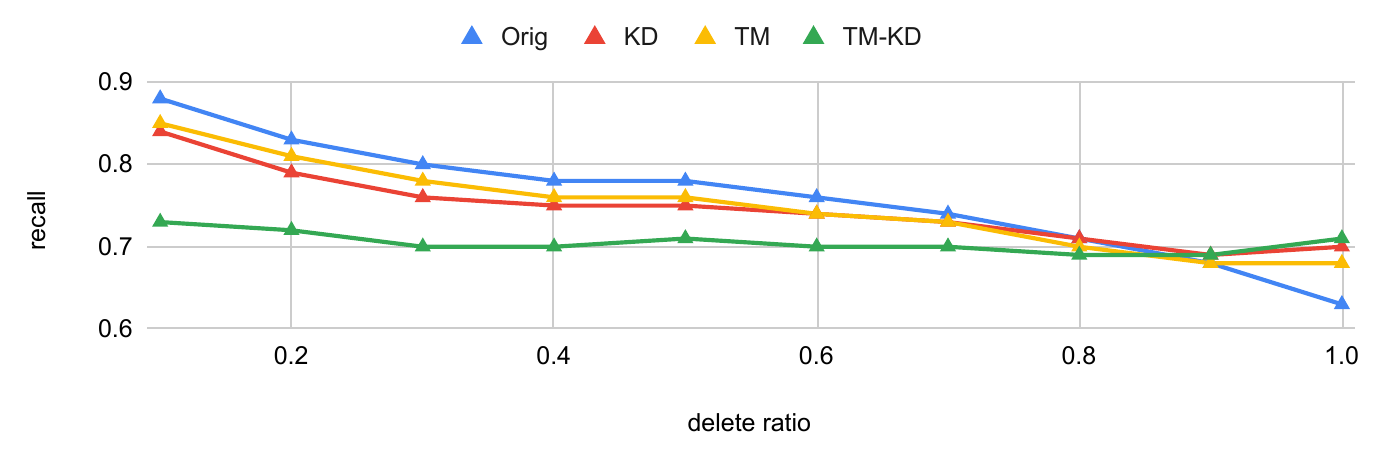}
  \caption{recall of random token}
  \label{fig:recall}
\end{subfigure}%
\caption{Precision and recall of random token prediction}
\label{fig:tok}
\end{figure}

\begin{figure}[ht]
\centering
\begin{subfigure}{.96\columnwidth}
  \centering
  \includegraphics[width=\columnwidth]{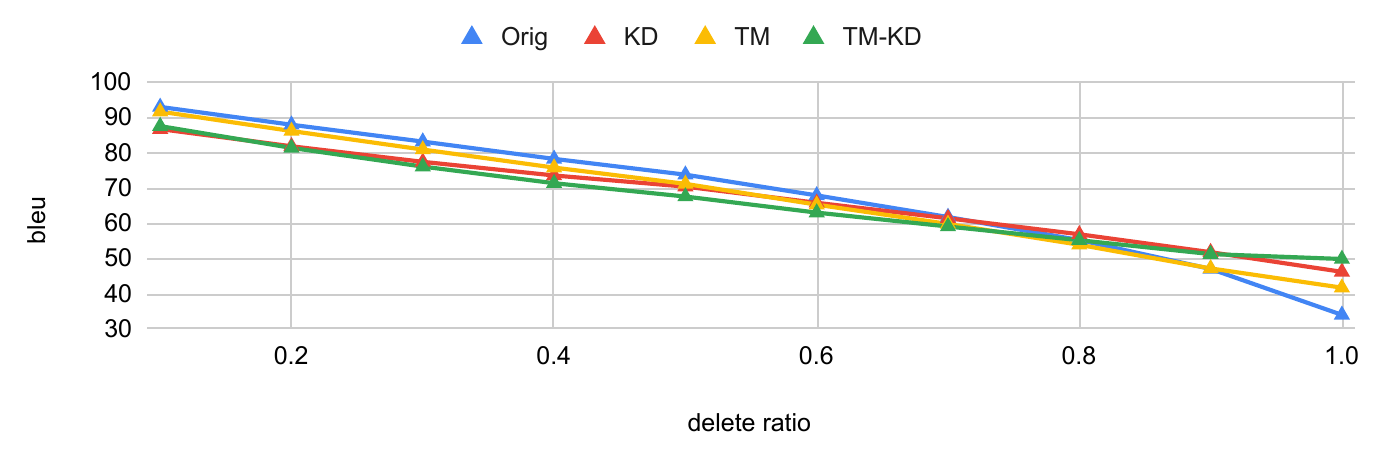}
  \caption{for random token prediction after one iteration}
  \label{fig:bleu-first}
\end{subfigure}%
\vskip\baselineskip
\begin{subfigure}{.96\columnwidth}
  \centering
  \includegraphics[width=\columnwidth]{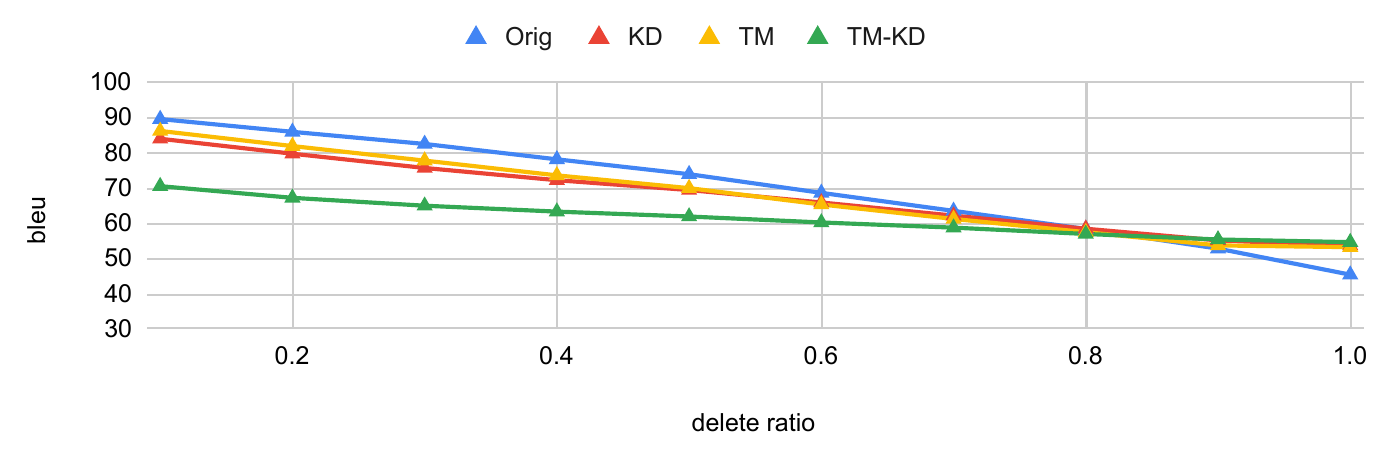}
  \caption{for random token prediction in the end}
  \label{fig:bleu-last}
\end{subfigure}%
\caption{Word level bleu score of random token prediction.}
\label{fig:tok-bleu}
\end{figure}

\paragraph{Observation} (1) From Figure~\ref{fig:pld}, predicting the length of several missing subwords is harder than complete words. (2) From Figure~\ref{fig:sub/word}, all models already achieve quite good token prediction results at first iteration; Orig-LevT performs badly when missing many subword tokens, but quite well at predicting word tokens. (3) From Figure~\ref{fig:tok}, Orig-LevT performs badly when decoding from scratch.

\section{Analysis of deletion module}
\subsection{Give no accuracy / no fluency decoding initialization}
From previous analysis, we find that the first generation is crucial during refinement, but how is model's ability to detect wrong translations and correct them? Here, we set two extreme decoding initialization situations: no accuracy and no fluency. 

'No accuracy' means that the initial sentence is grammatically correct but has nothing to do with the source sentence. We produce it by randomly reorder all reference target sentences and break the alignment between source and target. 'No fluency' means that the initial sentence is grammatically incorrect but contains all corresponding words. We produce it by randomly reorder words in the reference target sentence before BPE.

\begin{table}[ht]
\centering
\resizebox{\columnwidth}{!}{
\begin{tabular}{ccccc}
\hline
\textbf{}           & \textbf{Orig}                                                                    & \textbf{KD}                                                                      & \textbf{TM}                                                                         & \textbf{TM-KD}                                                                       \\ \hline
\textbf{iter num}   & 0.82                                                                             & 1.47                                                                             & 1.89                                                                                & 1.43                                                                                 \\
\textbf{pld 1 len}  & 19.75                                                                            & 19.75                                                                            & 19.75                                                                               & 19.75                                                                                \\
\textbf{del 1 len}  & 19.48                                                                            & 18.38                                                                            & 3.63                                                                                & 1.57                                                                                 \\
\textbf{pld 2 len}  & 20.18                                                                            & 21.73                                                                            & 17.17                                                                               & 18.62                                                                                \\
\textbf{final len}  & 21.35                                                                            & 23.05                                                                            & 18.52                                                                               & 19.82                                                                                \\
\textbf{final bleu} & \begin{tabular}[c]{@{}c@{}}2.1299 \\ 11.7/2.1/1.1/0.8 \\ BP = 1.000\end{tabular} & \begin{tabular}[c]{@{}c@{}}7.8504 \\ 19.7/8.2/5.6/4.2 \\ BP = 1.000\end{tabular} & \begin{tabular}[c]{@{}c@{}}47.5594\\ 70.7/56.6/48.2/42.0 \\ BP = 0.892\end{tabular} & \begin{tabular}[c]{@{}c@{}}53.9928 \\ 73.7/59.7/51.2/44.8 \\ BP = 0.957\end{tabular} \\ \hline
\end{tabular}}
\caption{No accuracy decoding initialization for four models.}
\label{tab:no-accuracy}
\end{table}

\begin{table}[ht]
\centering
\resizebox{\columnwidth}{!}{
\begin{tabular}{ccccc}
\hline
\textbf{}           & \textbf{Orig}                                                                        & \textbf{KD}                                                                          & \textbf{TM}                                                                          & \textbf{TM-KD}                                                                       \\ \hline
\textbf{iter num}   & 2.15                                                                                 & 1.88                                                                                 & 3.02                                                                                 & 1.72                                                                                 \\
\textbf{pld 1 len}  & 19.75                                                                                & 19.75                                                                                & 19.75                                                                                & 19.75                                                                                \\
\textbf{del 1 len}  & 7.14                                                                                 & 7.2                                                                                  & 19.2                                                                                 & 6.91                                                                                 \\
\textbf{pld 2 len}  & 17.33                                                                                & 19.79                                                                                & 40.74                                                                                & 21.24                                                                                \\
\textbf{final len}  & 18.59                                                                                & 21.2                                                                                 & 28.6                                                                                 & 20.31                                                                                \\
\textbf{final bleu} & \begin{tabular}[c]{@{}c@{}}40.2056 \\ 76.4/50.3/38.8/31.6 \\ BP = 0.863\end{tabular} & \begin{tabular}[c]{@{}c@{}}45.2821 \\ 69.6/50.0/39.2/31.5 \\ BP = 0.995\end{tabular} & \begin{tabular}[c]{@{}c@{}}30.3249 \\ 58.0/34.9/24.1/17.3 \\ BP = 1.000\end{tabular} & \begin{tabular}[c]{@{}c@{}}54.5834 \\ 73.5/59.1/50.3/43.7 \\ BP = 0.982\end{tabular} \\ \hline
\end{tabular}}
\caption{No fluency decoding initialization for four models.}
\label{tab:no-fluency}
\end{table}

\paragraph{Observation} (1) From Table~\ref{tab:no-accuracy}, we find that the delete module of Orig-LevT and KD-LevT does not look at the source sentence, but is more like a pure language model. Tm-LevT does look at the source. (2) From Table~\ref{tab:no-fluency}, all models can find grammatical errors, but the final bleu score is not as good as decoding from scratch. The current LevT decoder cannot make good use of 'correct but wrongly placed' words. 

\subsection{Try to delete more after first token insertion}
From Table~\ref{tab:no-accuracy}, we see that if the model does not delete much, there is no space to generate. Here, we try to force the model to delete more (or less) after first token insertion, then refine as usual. Note that the deletion prediction is a classification task, model gives scores of 'delete' and 'not delete', then chooses the higher one.

We perform $softmax$ on the two scores to make all results between $0$ and $1$, then set a threshold of deletion. If the predicted score is higher than the threshold, we decide to delete.

\begin{figure}[ht]
\centering
\begin{subfigure}{.48\columnwidth}
  \centering
  \includegraphics[width=\columnwidth]{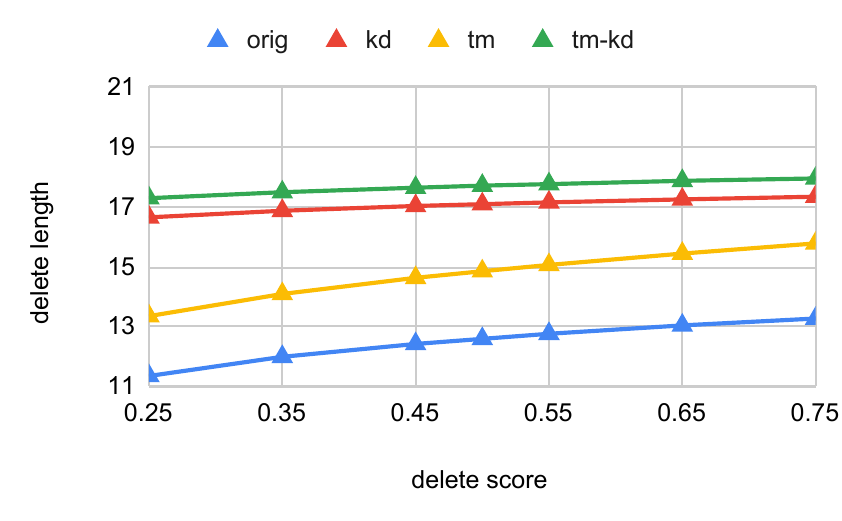}
  \caption{length after deletion}
  \label{fig:del-1}
\end{subfigure}%
\begin{subfigure}{.48\columnwidth}
  \centering
  \includegraphics[width=\columnwidth]{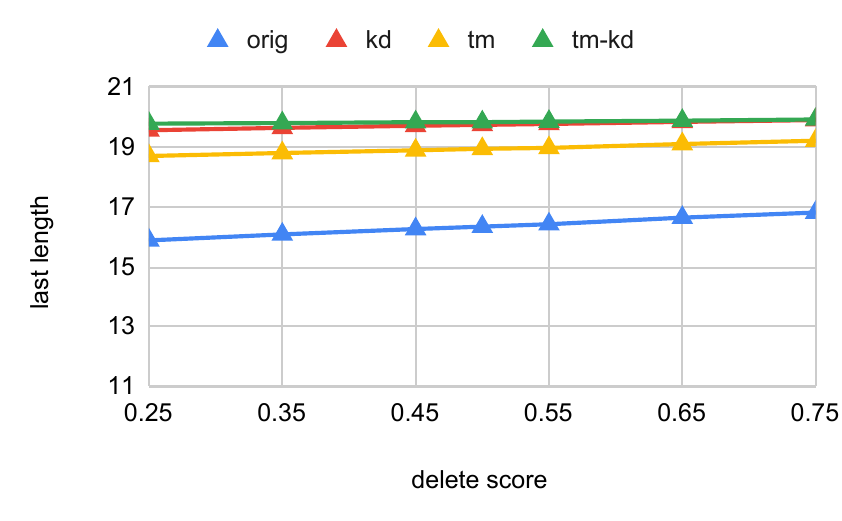}
  \caption{final length}
  \label{fig:del-last}
\end{subfigure}
\caption{Length of sentences for different deletion threshold.}
\label{fig:del}
\end{figure}

\paragraph{Observation} From Figure~\ref{fig:del}, (1) All models are quite confident of its prediction. The difference between threshold $0.25$ and threshold $0.75$ is only $2$. (2) KD models are more confident on deletion and have stable length.

\section{Conclusion}
In this report, we analyse decoding results of four models: original LevT, knowledge-distilled LevT, LevT with translation memory, and KD-LevT with translation memory. 

In terms of length, original LevT produces too short sentences due to its bad length prediction in the first iteration. An external length predictor can help, for example simply apply a ratio to source sentence length achieves good result.

In terms of subword and complete word generation, all four models do not generate enough subword. Original LevT is especially bad at predicting subword tokens, but good at generating complete words. It also performs bad when decoding from scratch.

In terms of deletion module, all models are confident on their deletion decision, especially KD models. However, the deletion module of original LevT and KD-LevT only plays the role of a language model and does not look at the source. LevT with translation memory does not have this problem.

Based on these observations, there are several possible ideas of improvement: add a length predictor, especially in the first iteration; train the model on more start-from-scratch cases; use another preprocessing method for subword; improve the cooperation between deletion prediction module and token prediction module, for example by a better training method.

\bibliography{anthology,custom}
\bibliographystyle{acl_natbib}




\end{document}